\documentclass[runningheads]{llncs}

\usepackage{custom}

\begin{document}
\maketitle              % typeset the header of the contribution
\begin{abstract}
Accurate segmentation is essential for echocardiography-based assessment of cardiovascular diseases (CVDs).
However, the variability among sonographers and the inherent challenges of ultrasound images hinder precise segmentation. 
By leveraging the joint representation of image and text modalities, Vision-Language Segmentation Models (VLSMs) can incorporate rich contextual information, potentially aiding in accurate and explainable segmentation.
However, the lack of readily available data in echocardiography hampers the training of VLSMs. 
In this study, we explore using synthetic datasets from Semantic Diffusion Models (SDMs) to enhance VLSMs for echocardiography segmentation.
We evaluate results for two popular VLSMs (CLIPSeg and CRIS) using seven different kinds of language prompts derived from several attributes, automatically extracted from echocardiography images, segmentation masks, and their metadata.
Our results show improved metrics and faster convergence when pretraining VLSMs on SDM-generated synthetic images before finetuning on real images.
The code, configs, and prompts are available at \url{https://github.com/naamiinepal/synthetic-boost}.

\keywords{Vision-Language Models \and Vision-Language Segmentation Models \and Echocardiography \and Synthetic Data}
\end{abstract}

\section{Introduction}
\label{sec:introduction}

Echocardiography (heart ultrasound) is an integral diagnostic tool for several cardiovascular diseases (CVDs).
It is widely used because it is cheap, portable, has no harmful radiation, and has a high temporal resolution (the ability to see high-definition images in real-time).
Accurately estimating clinically relevant quantitative measures in echocardiography images, such as cardiac substructure volumes and Ejection Fraction (EF), requires reliable segmentation algorithms.
However, segmenting various parts of the heart is challenging as the same standard plane image can have diverse appearances depending on the operator, and the presence of shadows, speckles, strong attenuation, and low contrast difference among areas of interest in ultrasound images \cite{avola2021ultrasound}.
Different CNN- and ViT-based \cite{dosovitskiy2020image} U-Net-like models \cite{deng2021transbridge,hatamizadeh2022unetr,isensee2021nnu,ronneberger2015u} are the state-of-the-art segmentation models that rely on supervised training with a relatively large set of annotated echocardiography images.
These segmentation models, however, must be trained on predefined classes that necessitate retraining or architecture changes (in the final layer) when new classes are required.
It is also challenging to manually intervene in or inject specific conditioning and make them explicitly benefit from the spatiotemporal relationships of different foreground structures.
Besides, they lack explainability and are not resilient to distribution shifts.

Recently, Vision-Language Models (VLMs) have been proposed that learn a joint representation of image and language \cite{furst2022cloob,huang2020pixel,jia2021scaling,li2021supervision,radford2021learning,singh2022flava,zhai2022lit}.
VLMs extract rich supplementary information via image and language prompt pairs, potentially aiding deep learning models to benefit from the richer information.
VLMs have one encoder each for image and language inputs, and the encoders are trained together to optimize a joint representation using losses such as contrastive loss. 
Vision-Language Segmentation Models (VLSMs) are adapted from VLMs where a decoder is added and trained on top of pretrained VLMs to segment the input image while leveraging information provided by language prompts \cite{luddecke2022image,rao2022denseclip,wang2022cris}.
However, almost all VLMs are trained using a large set of natural images, and no VLSMs are trained on an extensive collection of ultrasound datasets.
Although some recent methods show that VLMs and VLSMs could be finetuned on limited medical data \cite{qin2022medical}, the performance of these VLSMs is still below the supervised segmentation networks trained and optimized for specific datasets and foreground masks.

One major challenge to improving VLSMs for ultrasound images is the lack of large language-image paired datasets.
To address the limited data problem, generative models like GANs \cite{goodfellow2014generative} and diffusion models \cite{ho2020denoising} could generate images with a distribution closer to the real-world samples.
Stojanovski et al. \cite{stojanovski2023echo} trained Semantic Diffusion Models (SDMs) \cite{wang2022semantic} on the CAMUS dataset \cite{leclerc2019deep} to generate synthetic cardiac ultrasound images and showed that the segmentation model trained exclusively on a generated dataset results in a test dice score of $89.0 \pm 2.5$ in the CAMUS dataset.
The use of synthetic images has not been explored for VLSMs.
In this work, we explore whether the synthetic images from SDMs can improve the performance of VLSMs in echocardiography images.

Our primary contributions are as follows.

\begin{enumerate}
    \item We show that the VLSMs, pretrained on natural images, generalize to the real dataset (CAMUS) when finetuned on SDM-generated echocardiography images.
    \item We show that although numerous synthetic samples alone are not as good as a small number of real annotated data, the model finetuned on synthetic data is a good starting point for VLSMs to further finetune on real datasets.
\end{enumerate}

\section{Methodology}
\label{sec:methodology}

\subsection{Vision-Language Segmentation Models (VLSMs)}
\label{sec:vlsm}

\begin{figure}[t]
    \centering
    \includegraphics[width=\linewidth]{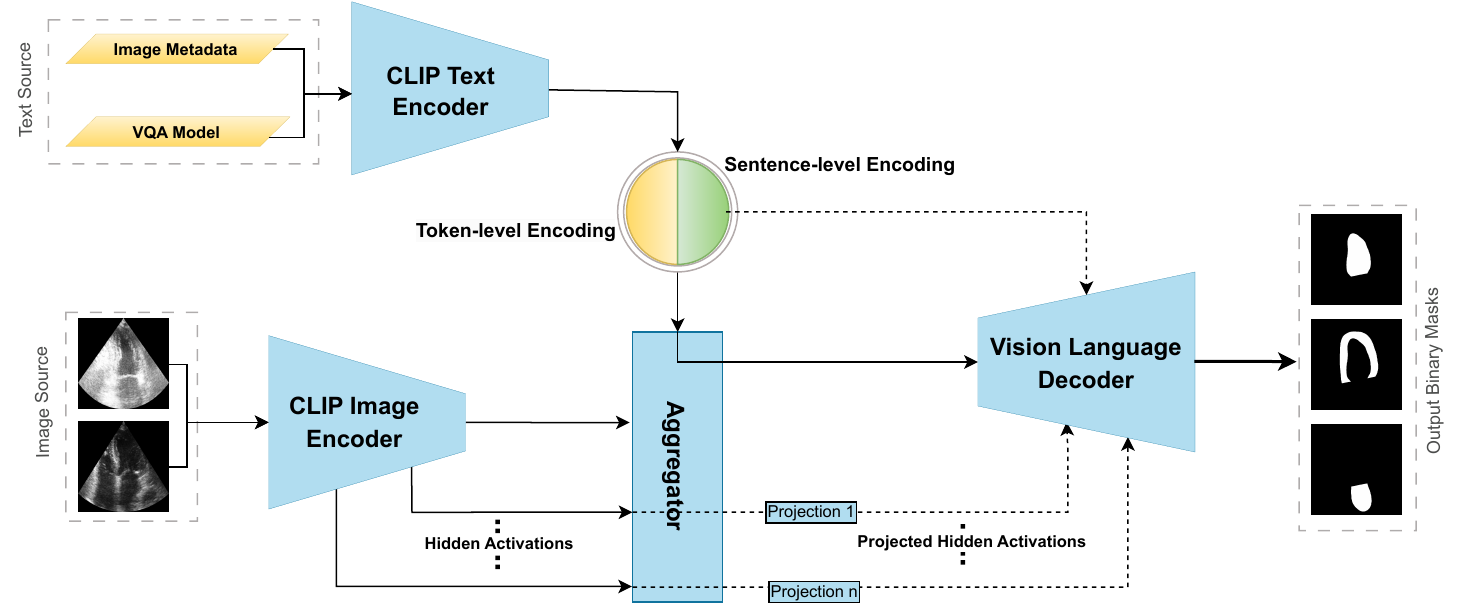}
    \caption{
        The basic architecture of CRIS and CLIPSeg VLSMs.
        The key components in the architecture are a \textit{Text Encoder}, an \textit{Image Encoder}, a \textit{Vision-Language Decoder (VLD)}, and an \textit{Aggregator}.
        The images and the corresponding prompts are passed to the CLIP image and text encoders, respectively. 
        The Aggregator generates intermediate representations utilizing image-level, sentence-level, or word-level representations to feed to the VLD.
        The VLD outputs a binary segmentation mask for an image-text pair.
    }
    \label{fig: architecture}
\end{figure}

CLIP \cite{radford2021learning} is a widely used VLM that jointly trains an image encoder and a text encoder to project semantically similar image-text pairs closer together and semantically disjoint image-text pairs farther apart.
As shown in \cref{fig: architecture}, the contrastive feature representation obtained from the two encoders of CLIP is fed to a vision-language decoder which generates a binary segmentation mask.
We investigate CLIPSeg \cite{luddecke2022image} and CRIS \cite{wang2022cris}, two state-of-the-art VLSMs for natural images, with various combinations of language prompts, real images, and synthetic images.

We use the publicly accessible CLIPSeg and CRIS weights learned during pretraining on natural image-text pairings \cite{kazemzadeh2014referitgame,wu2020phrasecut}.
The two VLSMs are finetuned on echocardiography datasets, starting from their publicly available pretrained weights.
The two echocardiography datasets are: (\textbf{i}) CAMUS \cite{leclerc2019deep}, and (\textbf{ii}) SDM CAMUS \cite{stojanovski2023echo}.
To test if pretraining with synthetic data boosts the segmentation performance on the real data, natural images pretrained VLSMs are further trained on extensive synthetic data and then finetuned on a smaller CAMUS dataset.

\subsection{Datasets}
\label{sec:datasets}

\subsubsection{CAMUS}

CAMUS \cite{leclerc2019deep} is a cardiac segmentation dataset containing 2D apical two-chamber (2C) and four-chamber (4C) views from $500$ patients at both end-diastole (ED) and end-systole (ES) cycles.
The dataset contains the semantic segmentation of the left ventricular cavity, the myocardium, and the left atrial cavity.
The original dataset randomly sampled images from $50$ patients as the official test split, and the remaining $450$ kept in the train split.
From those remaining $450$ patients, like Stojanovski et al. \cite{stojanovski2023echo}, we selected the first $50$ patients for validation and the remaining $400$ for the training.
The number of train/val/test images is $1,600$/$400$/$200$.

\subsubsection{Synthetic Echocardiography}

We use the synthetic echocardiography images proposed by Stojanovski et al. \cite{stojanovski2023echo}, generated using SDMs \cite{wang2022semantic}.
This model takes perturbed anatomical masks as conditioning information to denoise the noisy images and generates echocardiographic images.
Our experiments use $9,000$ synthetic images ($8,000$ for training and $1,000$ for validation) provided by the authors, the same splits they used to train and validate a vanilla U-Net \cite{ronneberger2015u} model.

\subsection{Prompt Engineering}
\label{sec:prompt_eng}

\subsubsection{Prompts for CAMUS}
\label{sec:prompt_eng_camus}

For our experiments, prompts, images, and masks are needed as triplets but are unavailable in the CAMUS dataset.
Finding the best prompt for the task is challenging, and creating the prompts manually for each image and mask pair is tedious and not scalable when the dataset size increases.
Also, the choice of prompts seems to significantly affect the performance of the VLMs in the medical domain \cite{qin2022medical}.

\begin{table}[b]
    \setlength{\tabcolsep}{5pt}
    \centering
    \caption{The description of the attribute and its possible values.
    The prompt number aside shows the prompt in which the attribute is introduced.}
    \label{tab:prompt_attributes}
    \begin{tabular}{l|ll}
         & \textbf{Description} & \textbf{Possible Values} \\
         \hline
         \textbf{P0} & Empty String \\
         \textbf{P1} & Target Structure & left ventricular cavity, myocardium, or left atrium cavity \\
         \textbf{P2} & Apical View & two-chamber view or four-chamber view \\
         \textbf{P3} & Cardiac Cycle & end of systole or diastole cycle \\
         \textbf{P4} & Patient's Sex & male or female \\
         \textbf{P5} & Patient's Age & all ages \\
         \textbf{P6} & Image Quality & good, medium, or poor \\
         \textbf{P7} & Structure's Shape & circle, triangle, oval, square, or rectangle \\
    \end{tabular}
\end{table}

We follow Poudel et al. \cite{poudel2023exploring} to generate automatic prompts adapted for the CAMUS dataset to explore if specific image features could be aligned to language prompts explaining those features.
The foreground cardiac structure's size and shape depend on the subjects' age, sex, and cardiac cycle phase.
Similarly, image quality information may help models adapt accordingly. 
As shown in \cref{tab:prompt_attributes}, various language prompts are designed by including words corresponding to the target structure name, its shape, the information about apical views, cardiac cycle phase, the subject's sex, the subject's age, and image quality (labeled by an expert within the CAMUS dataset).
There are $7$ attributes generated for the CAMUS dataset and $7$ prompts (\textbf{P1 - P7}) from the attributes, each added incrementally.
\textbf{P0} is an empty string.
The attributes in \textbf{P1 - P7} are ordered in descending order of the attribute's perceived importance (\textbf{P1} being the most important).

The sources of the attributes are listed below.

\begin{enumerate}
    \item \textbf{Image Filename}: We parse the images' filenames and masks to get the anatomical structure to segment, apical view, and cardiac cycle.
    \item \textbf{Image Metadata}: We parse the official metadata provided with the images and masks to get patients' sex, age, and image quality.
    \item \textbf{VQA Model}: We use OFA (One For All) VQA \cite{wang2022ofa} to get target structures' shapes.
    The VQA model is presented with the question, \textit{What is the shape of the \texttt{<structure>} in the green box?}.
    Here, the green box is the boundary of the target structure extracted from its mask.
\end{enumerate}

One example prompt \textbf{P7} with seven attributes: \textit{\textbf{Left ventricular cavity} of \textbf{oval shape} in \textbf{two-chamber view} in the cardiac ultrasound at the end of the \textbf{diastole cycle} of a \textbf{40-year-old female} with \textbf{poor image quality}.}

\subsubsection{Prompts for SDM CAMUS}
\label{sec:prompt_eng_sdm}

We did not use the image quality attribute in SDM CAMUS dataset as the synthetic images' quality is not annotated.
When synthesizing the prompts, we used the SDM CAMUS dataset's values derived from the original dataset for all other attributes: patient id, view information, and cardiac cycle.
One example prompt \textbf{P6} for the SDM CAMUS dataset: \textit{\textbf{Left ventricular cavity} of \textbf{oval shape} in \textbf{two-chamber view} in the cardiac ultrasound at the end of the \textbf{diastole cycle} of a \textbf{40-year-old female}.}

\section{Experimental Settings}
\label{sec:experimental_settings}

Unless specified, the VLSM's hyperparameters are the same as mentioned in the original implementation by the respective authors for all experiments.
The models are finetuned and inferred in NVIDIA GeForce RTX 3090, Titan Xp, and V100 GPUs.
We use float-16 mixed-precision training for models with different batch sizes of $32$ and $128$ for CRIS and CLIPSeg, respectively.
The batch sizes were chosen to utilize the full memory of the GPUs (maximum 24GB); since CRIS has a greater memory footprint than CLIPSeg, we reduced the former's batch size.

We use AdamW \cite{loshchilov2018decoupled} optimizer with the weight decay of $10^{-3}$ and an initial learning rate of $2 \times 10^{-3}$ and $2 \times 10^{-5}$ for CLIPSeg and CRIS, respectively.
The learning rate is reduced by a factor of $10$ if validation loss does not decrease for $5$ consecutive epochs\footnote{\url{https://pytorch.org/docs/stable/generated/torch.optim.lr_scheduler.ReduceLROnPlateau.html}}.

Three different strategies are employed to train the models: (\textbf{i}) training only on real data from the CAMUS dataset (\textit{real}), (\textbf{ii}) training only on synthetic data generated from the SDM model (\textit{synthetic}), and (\textbf{iii}) first training the model on the synthetic data, then finetuning on the real data (\textit{synth-PT:real-FT}).
CLIPSeg and CRIS resize the input images to $416 \times 416$ and $352 \times 352$, respectively. 
We normalize the resized images with the means and standard deviations provided by the respective models.
No augmentation and further post-processing are done to assess the models' raw performance.

We used the weighted sum of soft Dice and Binary Cross Entropy losses with weights $1$ and $0.2$, respectively.
All the dice scores are computed at $512 \times 512$ (nearly the median width of the dataset), resizing the model's output when required.
For each experiment, the metrics reported are for the model with the best dice score on the validation set, across the epochs, with an output threshold of $0.5$ on the predicted binary segmentation map.

To study the ability of the VLSMs to represent the alignment of image-text pairs, we perform two experiments: (\textbf{i}) freezing the VLM encoders of CRIS and CLIPSeg, and (\textbf{ii}) unfreezing the VLM encoders during finetuning on all datasets.
The dice score for the unfrozen encoders is shown in the \cref{tab:combined_dice} whereas that of the frozen ones is demonstrated in \cref{tab:combined_dice_frozen}.

\section{Results}
\label{sec: results}

\subsection{Synthetic data is better than no data}

\cref{tab:combined_dice} shows that while the VLSMs pretrained on natural images perform very poorly on ultrasound images in zero-shot segmentation, models trained on synthetic data provide much better results in real ultrasound images.

\begin{table}[b]
    \centering
    \caption{The dice score (mean $\pm$ std) of models trained and validated using various strategies and evaluated on the CAMUS's official test split when the encoders of the VLMs are unfrozen.
    The zero-shot performance of the models is extracted from Poudel et al. \cite{poudel2023exploring} for comparison.}
    \label{tab:combined_dice}
    \resizebox{\linewidth}{!}{%
        \begin{tabular}{l|l|cccccccc}
            \multirow{2}{*}{\textbf{Strategy}} & \textbf{Prompt $\rightarrow$} & \multirow{2}{*}{\textbf{P0}} & \multirow{2}{*}{\textbf{P1}} & \multirow{2}{*}{\textbf{P2}} & \multirow{2}{*}{\textbf{P3}} & \multirow{2}{*}{\textbf{P4}} & \multirow{2}{*}{\textbf{P5}} & \multirow{2}{*}{\textbf{P6}} & \multirow{2}{*}{\textbf{P7}} \\
            \cline{2-2}
            & \textbf{Model $\downarrow$} \\
            \hline
            \multirow{2}{*}{\em zeroshot} & \textbf{CLIPSeg} & $0.00 \smallStd{0.0}$ & $0.00 \smallStd{0.0}$ & $0.21 \smallStd{1.8}$ & $0.16 \smallStd{1.9}$ & $0.19 \smallStd{2.1}$ & $0.51 \smallStd{3.7}$ & $0.46 \smallStd{3.1}$ & $1.81 \smallStd{6.6}$ \\
            & \textbf{CRIS} & $23.53  \smallStd{12.0}$ & $9.04  \smallStd{13.9}$ & $8.36  \smallStd{13.2}$ & $8.24  \smallStd{13.2}$ & $8.24  \smallStd{13.2}$ & $8.24  \smallStd{13.2}$ & $8.24  \smallStd{13.2}$ & $5.45 \smallStd{10.4}$ \\
            \hline
            \multirow{2}{*}{\em synthetic} & \textbf{CLIPSeg} & $45.69 \smallStd{13.2}$ & $84.24 \smallStd{12.0}$ & $84.87 \smallStd{10.9}$ & $85.27 \smallStd{9.7}$ & $84.38 \smallStd{11.0}$ & $83.18 \smallStd{12.8}$ & $83.32 \smallStd{12.5}$ & N/A\\
            & \textbf{CRIS} & $42.29 \smallStd{17.6}$ & $84.72 \smallStd{11.9}$ & $84.72 \smallStd{10.5}$ & $85.48 \smallStd{10.2}$ & $85.12 \smallStd{11.2}$ & $85.84 \smallStd{10.0}$ & $84.35 \smallStd{13.3}$ & N/A\\
            \hline
            \multirow{2}{*}{\em real} & \textbf{CLIPSeg} & $\mathbf{46.52 \smallStd{13.3}}$ & $88.53 \smallStd{7.2}$ & $88.81 \smallStd{7.2}$ & $88.77 \smallStd{7.2}$ & $88.58 \smallStd{7.7}$ & $88.27 \smallStd{7.4}$ & $88.45 \smallStd{7.5}$ & $88.16 \smallStd{8.0}$ \\
            & \textbf{CRIS} & $46.46 \smallStd{13.1}$ & $91.00 \smallStd{6.3}$ & $91.03 \smallStd{6.2}$ & $89.9 \smallStd{7.6}$ & $90.94 \smallStd{6.6}$ & $90.87 \smallStd{6.4}$ & $90.79 \smallStd{7.1}$ & $90.99 \smallStd{6.3}$ \\
            \hline
            \multirow{2}{*}{\em \specialcell{synth-PT:\\real-FT}} & \textbf{CLIPSeg} & $46.26 \smallStd{13.2}$ & $88.56 \smallStd{7.5}$ & $89.44 \smallStd{6.9}$ & $89.8 \smallStd{6.8}$ & $88.68 \smallStd{7.5}$ & $88.55 \smallStd{7.4}$ & $89.36 \smallStd{6.8}$ & $89.53 \smallStd{6.6}$\\
            & \textbf{CRIS} & $41.09 \smallStd{18.6}$ & $\mathbf{91.26 \smallStd{6.1}}$ & $\mathbf{91.39 \smallStd{5.9}}$ & $\mathbf{91.12 \smallStd{6.3}}$ & $\mathbf{91.04 \smallStd{7.2}}$ & $\mathbf{91.23 \smallStd{6.4}}$ & $\mathbf{91.11 \smallStd{6.8}}$ & $\mathbf{91.08 \smallStd{6.6}}$
        \end{tabular}%
    }
\end{table}

\subsection{Real data is better than synthetic data}

\begin{figure}[t]
    \centering
    \includegraphics[width=\linewidth]{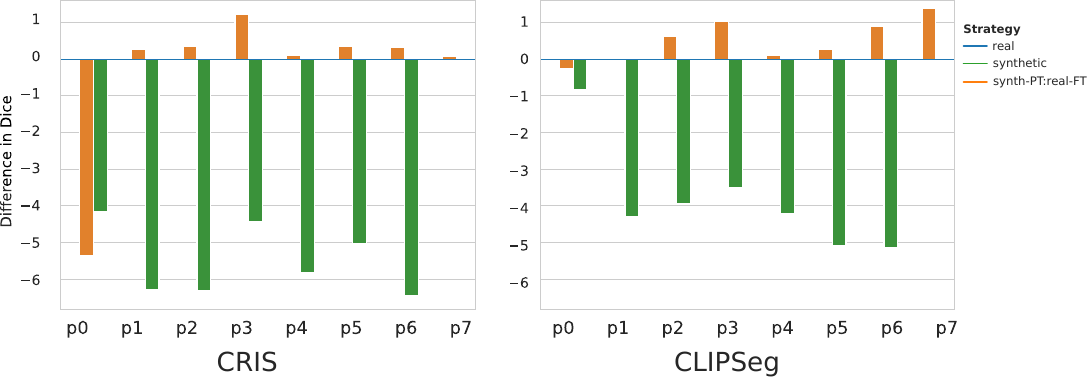}
    \caption{Difference in mean dice scores between different training strategies for CLIPSeg and CRIS for different prompts, relative to real.
    Pretraining on synthetic data before finetuning them on real data helps to improve the performance of VLSMs.}
    \label{fig:dice_camus_diff}
\end{figure}

\cref{fig:dice_camus_diff} shows that VLSMs have better dice scores when finetuned in real data than finetuning only in synthetic data.
When comparing the best dice scores for both strategies, the models trained on the synthetic dataset have a lower dice score ($-5.19$), which is statistically highly significant by the Wilcoxon signed-rank test \cite{wilcoxon1992individual} with a p-value of $8.8 \times 10^{-73}$, on the official test split of real images.

\subsection{Pretraining on synthetic data helps in finetuning on real data}

In both CRIS and CLIPSeg, the pretraining on synthetic data and then finetuning on real data (\textit{synth-PT:real-FT} strategy) performs better than the experiments trained with either real or artificial images as illustrated in \cref{fig:dice_camus_diff}.
This second stage pretraining strategy has a higher dice score ($+0.34$), which is statistically significant by the Wilcoxon signed-rank test \cite{wilcoxon1992individual} with a p-value of $8.3 \times 10^{-6}$, than the models that haven't seen synthetic data.

\subsection{Unfreezing VLM encoders during finetuning affects models differently}

For \textbf{P0} (empty prompt), the output class is ambiguous for the models.
From \cref{tab:combined_dice_frozen}, we can infer that CLIPSeg dealt with this obscurity by predicting a segmentation map of a union of all the classes, while CRIS chose just noise (all zeros in the case of the last strategy).

\begin{table}[b]
    \centering
    \caption{The dice score (mean $\pm$ std) on the CAMUS's official test split when the encoders of the VLMs are frozen.}
    \label{tab:combined_dice_frozen}
    \resizebox{\linewidth}{!}{%
        \begin{tabular}{l|l|cccccccc}
            \multirow{2}{*}{\textbf{Strategy}} & \textbf{Prompt $\rightarrow$} & \multirow{2}{*}{\textbf{P0}} & \multirow{2}{*}{\textbf{P1}} & \multirow{2}{*}{\textbf{P2}} & \multirow{2}{*}{\textbf{P3}} & \multirow{2}{*}{\textbf{P4}} & \multirow{2}{*}{\textbf{P5}} & \multirow{2}{*}{\textbf{P6}} & \multirow{2}{*}{\textbf{P7}} \\
            \cline{2-2}
            & \textbf{Model $\downarrow$} \\
            \hline
            \multirow{2}{*}{\em synthetic} & \textbf{CLIPSeg} & $45.71 \smallStd{13.6}$ & $84.08 \smallStd{11.1}$ & $84.01 \smallStd{10.5}$ & $84.48 \smallStd{11.0}$ & $84.02 \smallStd{11.0}$ & $84.47 \smallStd{10.7}$ & $85.47 \smallStd{9.3}$ & N/A \\
            & \textbf{CRIS} & $35.13 \smallStd{19.5}$ & $84.19 \smallStd{13.0}$ & $84.02 \smallStd{12.2}$ & $84.62 \smallStd{11.9}$ & $84.94 \smallStd{11.5}$ & $84.23 \smallStd{12.3}$ & $80.70 \smallStd{17.0}$ & N/A \\
            \hline
            \multirow{2}{*}{\em real} & \textbf{CLIPSeg} &  $\mathbf{46.52 \smallStd{13.2}}$ & $88.81 \smallStd{7.2}$ & $89.04 \smallStd{7.0}$ & $88.65 \smallStd{7.3}$ & $89.05 \smallStd{7.2}$ & $88.54 \smallStd{7.5}$ & $88.61 \smallStd{7.5}$ & $88.54 \smallStd{7.6}$ \\
            & \textbf{CRIS} & $26.84 \smallStd{16.2}$ & $88.41 \smallStd{8.7}$ & $88.71 \smallStd{8.6}$ & $88.62 \smallStd{8.8}$ & $88.55 \smallStd{8.7}$ & $88.48 \smallStd{8.6}$ & $88.85 \smallStd{8.4}$ & $88.40 \smallStd{9.8}$ \\
            \hline
            \multirow{2}{*}{\em \specialcell{synth-PT:\\real-FT}} & \textbf{CLIPSeg} & $46.5 \smallStd{13.3}$ & $89.07 \smallStd{7.1}$ & $89.09 \smallStd{7.1}$ & $89.24 \smallStd{6.7}$ & $89.24 \smallStd{6.9}$ & $88.91 \smallStd{7.2}$ & $\mathbf{89.12 \smallStd{7.0}}$ & $89.14 \smallStd{7.0}$ \\
            & \textbf{CRIS} & $0.04 \smallStd{0.5}$ & $\mathbf{89.21 \smallStd{7.9}}$ & $\mathbf{89.54 \smallStd{7.4}}$ & $\mathbf{89.26 \smallStd{7.5}}$ & $\mathbf{89.41 \smallStd{7.6}}$ & $\mathbf{89.34 \smallStd{7.8}}$ & $89.03 \smallStd{9.2}$ & $\mathbf{89.34 \smallStd{8.2}}$
        \end{tabular}%
    }
\end{table}

\begin{figure}[t]
    \centering
    \includegraphics[width=\linewidth]{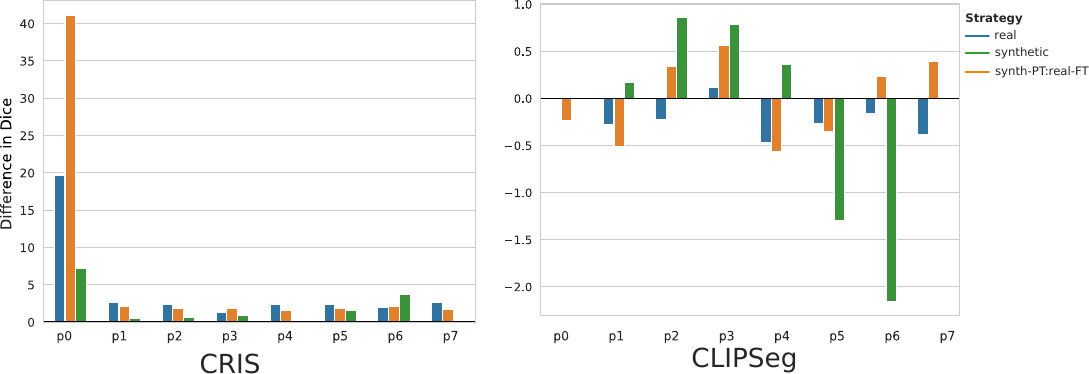}
    \caption{Difference between mean dice scores when the encoders are frozen and when the encoders are trained for different prompts.
    CRIS's model performance improves when the encoders are trained along with the decoder.
    In contrast, CLIPSeg's performance degrades when encoders are trained.
    }
    \label{fig:freeze-unfreeze-diff}
\end{figure}

\cref{fig:freeze-unfreeze-diff} shows that CRIS's performance improves when encoders are not frozen during finetuning.
In contrast, CLIPSeg's performance degrades when the encoders are unfrozen for the CAMUS dataset (real one), which seems to have improved when synthetic data is introduced.

\section{Discussion}
\label{sec:discussion}

Although the VLSMs do not improve over the state-of-the-art segmentation models on the CAMUS dataset (according to the leaderboard\footnote{\url{https://www.creatis.insa-lyon.fr/Challenge/camus/results.html}. Updated January 2023}, the maximum mean dice is $94.1$ \cite{ling2022reaching}), it is promising that they are close.
Pretraining with the synthetic samples followed by finetuning in real samples improves the results compared to finetuning on real examples without synthetic pretraining. 
One exciting direction to explore in the future is to train real and synthetic data together while indicating in the language prompt whether the sample is real or artificial.

The VLSMs pretrained on natural image-language pairs do not seem to have captured the language-image relationships common in ultrasound images.
Thus, when finetuning the encoders of VLSMs, the performance improved compared to freezing the encoders and finetuning only the decoder.
CRIS's performance is always better when the encoders are finetuned for every strategy, but CLIPSeg only performs better when the synthetic dataset is introduced.
Unfrozen CLIPSeg performing better when the dataset size is increased may be because, for CRIS, Wang et al. \cite{wang2022cris} finetuned the CLIP encoders and the vision-language decoder for the segmentation task, whereas, 
L{\"u}ddecke et al. \cite{luddecke2022image} froze the encoders for CLIPSeg.
Thus, CLIPSeg's encoder representation is likely not well adapted for segmentation as our finetuning of the encoder is limited to only a few thousand samples. 

SDM CAMUS \cite{stojanovski2023echo} is generated by applying random augmentations to the mask of the CAMUS dataset.
As the dataset was developed by utilizing all the labeled image-mask pairs in the training set, and the images could not be generated without the corresponding mask, this questions the ``synthetic'' portion of the method (or dataset).
This dataset does not solve the medical image segmentation's limited paired-data availability problem by generating new examples.
Instead, this is more akin to data augmentation, where the existing annotated set is augmented with a few thousand transformed pairs by perturbing existing masks and textures.
An important direction in the future would be to find ways to generate aligned synthetic triplets of language, image, and mask at scale without annotated image-mask pairs.

\section{Conclusion}

Recent VLSMs trained in large image-language pairs of natural photographs perform close to the state-of-the-art on the CAMUS echocardiography dataset when finetuned on the automatically generated prompts.
Augmenting training sets with synthetic images generated from SDM improves VLSMs' performance.
However, using a relatively large number of synthetic data alone is still inferior to using a relatively small number of real annotated data.
This suggests that more work is needed in generating better synthetic images whose distribution is closer to the real data distribution for the echocardiography dataset.
Nevertheless, the synthetic data finetuned model checkpoint seems to be a good starting point for the segmentation models to finetune on the real dataset, resulting in improved metrics and faster convergence (on an average, $4.55$ and $1.71$ times faster for CRIS and CLIPSeg, respectively).
While there is a significant potential for VLSMs for ultrasound image segmentation, there is a need to develop methods that can generate numerous consistent, realistic, but synthetic triplets of image, language, and segmentation masks if one wants to leverage the power of VLSMs.

\bibliographystyle{splncs04}
\bibliography{ms}

\end{document}